\PassOptionsToPackage{table}{xcolor} 
\documentclass[sigconf]{acmart}

\usepackage{enumitem}
\usepackage{multirow, tabularx}
\usepackage{booktabs} 
\usepackage{makecell}
\usepackage{subfigure}
\usepackage{amsmath}
\usepackage{arydshln} 
\usepackage{array} 
\usepackage[table]{xcolor} 
\usepackage{xspace} 
\usepackage{diagbox}

\makeatletter
\newcommand{\thickhline}{
	\noalign {\ifnum 0=`}\fi \hrule height 1pt
	\futurelet \reserved@a \@xhline
}
\makeatother

\newcommand{\pub}[1]{{\color{gray}{\tiny{[{#1}]}}}}
\definecolor{mygray}{gray}{.9}
\definecolor{mygreen}{RGB}{93,173,85}
\definecolor{myred}{RGB}{252,66,70}
\def\eg{\emph{e.g.}} 
\def\ie{\emph{i.e.}} 
\def\etal{\emph{et al.}}

\newcommand{\reshl}[2]{
	\textbf{#1} \fontsize{7.5pt}{1em}\selectfont\color{mygreen}{$\!\uparrow\!$ \textbf{#2}}
}
\newcommand{\reshll}[2]{
	#1 \fontsize{7.5pt}{1em}\selectfont\color{mygreen}{$\!\uparrow\!$ \textbf{#2}}
}

\AtBeginDocument{%
\providecommand\BibTeX{{%
		\normalfont B\kern-0.5em{\scshape i\kern-0.25em b}\kern-0.8em\TeX}}}

\setcopyright{acmlicensed}
\copyrightyear{2023}
\acmYear{2023}
\acmConference[SIGIR '23] {Proceedings of the 46th International ACM SIGIR Conference on Research and Development in Information Retrieval}{July 23--27, 2023}{Taipei, Taiwan.}
\acmBooktitle{Proceedings of the 46th International ACM SIGIR Conference on Research and Development in Information Retrieval (SIGIR '23), July 23--27, 2023, Taipei, Taiwan}
\acmPrice{15.00}
\acmISBN{978-1-4503-9408-6/23/07}
\acmDOI{10.1145/3539618.3591712}

\begin{document}


\title{Learnable Pillar-based Re-ranking for Image-Text Retrieval}


\author{Leigang Qu}
\email{leigangqu@gmail.com}
\affiliation{
  \institution{National University of Singapore}
  \country{}
}

\author{Meng Liu}
\email{mengliu.sdu@gmail.com}
\affiliation{
  \institution{Shandong Jianzhu University}
  \country{}
}

\author{Wenjie Wang}
\authornote{Corresponding author: Wenjie Wang (wenjiewang96@gmail.com).
This research is supported by the Defence Science and Technology Agency, the National Natural Science Foundation of China under Grant 62006142, and the Special Fund for distinguished professors of Shandong Jianzhu University.
}
\email{wenjiewang96@gmail.com}
\affiliation{
  \institution{National University of Singapore}
  \country{}
}

\author{Zhedong Zheng}
\email{zdzheng@nus.edu.sg}
\affiliation{
  \institution{National University of Singapore}
  \country{}
}

\author{Liqiang Nie}
\email{nieliqiang@gmail.com}
\affiliation{
  \institution{Harbin Institute of Technology (Shenzhen)}
  \country{}
}

\author{Tat-Seng Chua}
\email{dcscts@nus.edu.sg}
\affiliation{
  \institution{National University of Singapore}
  \country{}
}

\begin{abstract}
Image-text retrieval aims to bridge the modality gap and retrieve cross-modal content based on semantic similarities. 
Prior work usually focuses on the pairwise relations (\ie, whether a data sample matches another) but ignores the \textit{higher-order neighbor relations} (\ie, a matching structure among multiple data samples). 
Re-ranking, a popular post-processing practice, has revealed the superiority of capturing neighbor relations in single-modality retrieval tasks. However, it is ineffective to directly extend existing re-ranking algorithms to image-text retrieval. In this paper, we analyze the reason from four perspectives, \ie, generalization, flexibility, sparsity, and asymmetry, and propose a novel learnable pillar-based re-ranking paradigm. Concretely, we first select top-ranked intra- and inter-modal neighbors as \textit{pillars}, and then reconstruct data samples with the neighbor relations between them and the pillars. In this way, each sample can be mapped into a multimodal pillar space only using similarities, ensuring generalization. After that, we design a neighbor-aware graph reasoning module to flexibly exploit the relations and excavate the sparse positive items within a neighborhood. We also present a structure alignment constraint to promote cross-modal collaboration and align the asymmetric modalities. On top of various base backbones, we carry out extensive experiments on two benchmark datasets, \ie, Flickr30K and MS-COCO, demonstrating the effectiveness, superiority, generalization, and transferability of our proposed re-ranking paradigm.
\end{abstract}





\maketitle

\section{Introduction}\label{sec:introduction}
Image-text retrieval (ITR), as a fundamental vision-language task, has attracted substantial attention over the past decade~\cite{carvalho2018cross, qu2021dynamic, chen2021learning}.
It aims to bridge the heterogeneous modality gap and achieve semantical matching by bidirectional retrieval
: 1) image-to-text retrieval, searching for correct descriptive texts given an image, and 2) text-to-image retrieval, finding the relevant images for a given text. 
By pursuing cross-modal understanding and reasoning, ITR has become essential to other visual-language applications such as visual question answering~\cite{anderson2018bottom} and video moment retrieval~\cite{liu2018attentive}.

\begin{figure}[t]
	\includegraphics[width=0.46\textwidth]{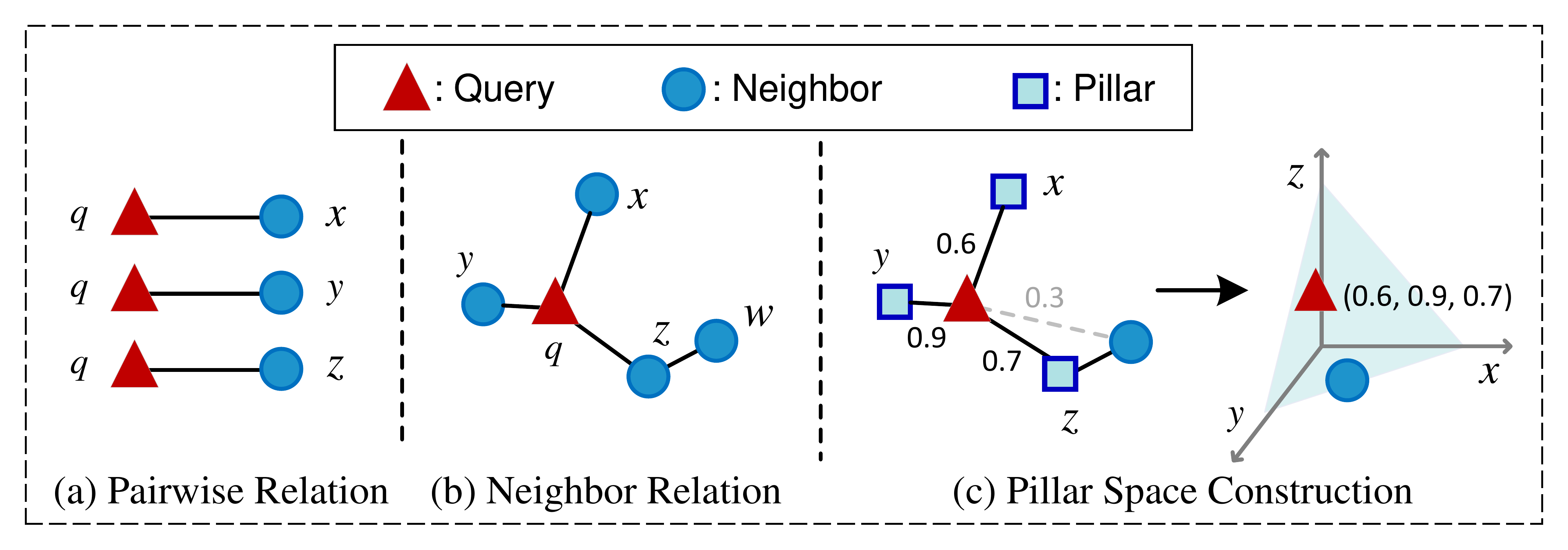}
	\vspace{-3ex}
	\caption{Illustration of (a) Pairwise Relation, (b) Neighbor Relation, and (c) Pillar Space Construction. Compared with the pairwise relations, the higher-order neighbor relation is a matching structure including the query-neighbor (\eg, $q \leftrightarrow x$) and neighbor-neighbor (\eg, $z \leftrightarrow w$) relations. In a neighborhood, top-ranked neighbors are selected as pillars to construct the pillar space to represent data samples. Numbers denote similarities calculated by a base backbone. }
\label{fig:neighbor_relations}
\vspace{-3ex}
\end{figure}

\begin{figure}[t]
	\includegraphics[width=0.32\textwidth]{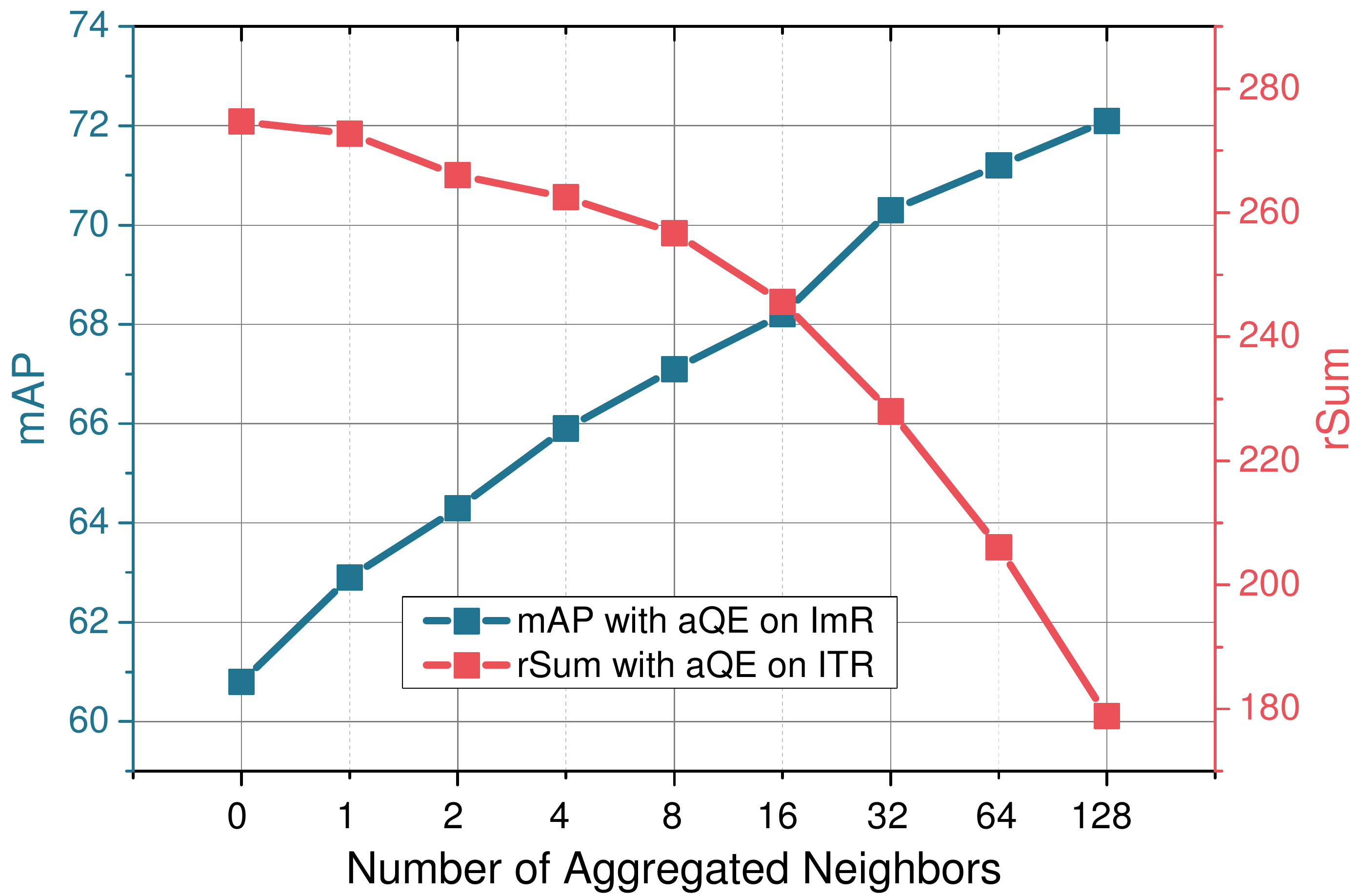}
	\vspace{-2ex}
	\caption{Retrieval performance comparison between Image Retrieval (ImR) and Image-Text Retrieval (ITR) using the classical re-ranking method $\alpha$QE~\cite{radenovic2018fine}. We use GeM (AP)~\cite{revaud2019learning} as the base backbone to evaluate ImR on $\mathcal{R}$Paris6K (Hard)~\cite{philbin2008lost}, and VSE$\infty$~\cite{chen2021learning} for ITR on Flickr30K~\cite{young2014image}. The results without re-ranking are shown when the neighbor number is 0. 
    }
\label{fig:ir_vs_itr}
\vspace{-3.5ex}
\end{figure}

Existing approaches to ITR fall into two categories:
The first category, \ie, \textit{two-tower} framework~\cite{li2019visual, chen2021learning}, separately maps the visual and textual samples into a joint embedding space to calculate the cross-modal similarity. 
In contrast, the second category, \ie, \textit{one-tower} framework~\cite{carvalho2018cross, qu2021dynamic}, models fine-grained interactions by the cross-attention mechanism and directly outputs the similarity. 
Despite impressive progress, both categories only consider the pairwise matching relations between an image and a text, as shown in Figure~\ref{fig:neighbor_relations}(a). They completely ignore the higher-order \textbf{neighbor relations} among multiple images and texts illustrated in Figure~\ref{fig:neighbor_relations}(b), and therefore suffer from suboptimal retrieval performance.
As a prevailing post-processing technology to explore neighbor relations, re-ranking has revealed remarkable effectiveness in conventional single-modality retrieval tasks, such as image retrieval~\cite{iscen2017efficient,zheng2020vehiclenet}, document retrieval~\cite{pang2020setrank, matsubara2020reranking}, and recommendation~\cite{xi2022multi}. As shown by the blue line in Figure~\ref{fig:ir_vs_itr}, the retrieval performance of image retrieval can be largely improved via query expansion re-ranking (\ie, $\alpha$QE~\cite{radenovic2018fine}). 
This motivates us to investigate re-ranking to exploit neighbor relations in the context of ITR. 

In general, existing re-ranking approaches can be divided into three main categories: Query Expansion (QE)~\cite{chum2007total, radenovic2018fine}, Diffusion~\cite{iscen2017efficient}, and Neighbor-based methods~\cite{zhong2017re}. 
However, they are not suitable for cross-modal ITR due to the following challenges: 
1) \textbf{Architecture Generalization}. Some re-ranking methods strongly rely on specific model architectures. For example, QE can be only applied to two-tower frameworks since it requires intermediate features for expansion~\cite{radenovic2018fine}, and thus can not well serve the one-tower framework. 
2) \textbf{Representation Flexibility}. Existing re-ranking approaches~\cite{radenovic2018fine, iscen2017efficient, zhong2017re}
are designed with rigid or restricted rules from prior knowledge, making them difficult to automatically explore neighbor relations in complex multimodal data.
Specifically, QE aggregates neighbors to enrich the query in a limited unidirectional way, the diffusion mechanism performs ranking on fixed manifolds, and neighbor-based methods~\cite{zhong2017re, qin2011hello}
heavily depend on certain pre-defined neighbor relations.
3) \textbf{Sparse Relevance}. The average ratio of positive items over the benchmark datasets of ITR (\eg, MS-COCO (5K)~\cite{lin2014microsoft}) is  much smaller than those of image retrieval (\eg, $\mathcal{R}$Paris6K~\cite{philbin2008lost}), specifically 0.02\% vs. 2.3\%. 
This makes re-ranking on ITR more challenging since noisy negative information are widely propagated among items when capturing the higher-order neighbor relations. 
And 4) \textbf{Modality Asymmetry}. 
The two asymmetry retrieval directions of ITR result in different similarity distributions and heterogeneous neighbors, yet existing re-ranking methods are designed for single-modality retrieval, thus failing to consider the asymmetry problem.
As shown in Figure~\ref{fig:ir_vs_itr}, we can observe the great performance degradation of $\alpha$QE (the red line) as the aggregated neighbors increase, even though we choose a two-tower backbone to fit it. We argue that it is caused by flexibility, sparsity, and asymmetry challenges.

To handle the above challenges, we present a \textbf{Lea}rnable \textbf{P}illar-based \textbf{R}e-\textbf{R}anking (\textbf{LeaPRR}) framework for ITR. 
1) First, to well fit in with one-tower and two-tower frameworks, we abandon any intermediate content features and only take the final similarities which can be calculated by any framework as the input to our re-ranking model. 
For the sake of informative entity (\ie, queries and items) representation, we define top-ranked multimodal neighbors as \textit{pillars} and then reconstruct entities with the similarities between pillars and these entities, as shown in Figure~\ref{fig:neighbor_relations}(c). In this way, each entity can be mapped into a pillar space. Different from the conventional modality-independent content space, the pillar space is constructed with multimodal pillars to alleviate the asymmetry problem.
2) Moreover, as for the flexibility and sparsity challenges, we propose a neighbor-aware graph reasoning module based on the neighborhood centered with a given query, to flexibly explore complicated neighbor relations, including the query-neighbor and neighbor-neighbor ones. 
3) To further deal with the high sparsity of positive items, we exert local and global ranking constraints to reinforce the discriminative power of refined pillar features, which can adaptively absorb valuable information and resist noise from neighbors.
4) Finally, we present a structure alignment constraint by transforming contextual structures from one modality to another, further combating the asymmetry issue. 

To sum up, our main contributions are as follows:
\begin{itemize}[leftmargin=*]
\item We discuss the limitations of existing re-ranking approaches in the ITR task and present four challenges, \ie, generalization, flexibility, sparsity, and asymmetry. 

\item To deal with the four challenges, we reformulate re-ranking in the multi-modality field and propose a learnable pillar-based re-ranking paradigm for the first time, which is plug-and-play and applicable to all existing ITR backbones. 

\item Extensive experiments based on different base ITR backbones on two datasets, \ie, Flickr30K~\cite{young2014image} and MS-COCO~\cite{lin2014microsoft}, validate the effectiveness and superiority of our model. The codes and settings are released at \url{https://github.com/LgQu/LeaPRR}.
\end{itemize}

\section{Related Work}
\subsection{Image-Text Retrieval}
Current ITR models can be divided into two categories, \ie, \textit{two-tower framework} and \textit{one-tower framework}, in accordance with the manner for modality interaction. 
As for the two-tower framework~\cite{frome2013devise, faghri2017vse++, zheng2020dual, qu2020context, chen2021learning}, images and texts are independently mapped into a joint feature space in which the semantic similarities are calculated via cosine function or Euclidean distance. For example, Frome \etal~\cite{frome2013devise} represented visual and textual instances in a common modality-agnostic space to assess semantic relevance. 
To further improve the discriminative power, Zheng \etal~\cite{zheng2020dual} argued that the commonly used ranking loss is not effective confronted with large-scale multimodality data and presented a new instance loss to exploit the intra-modal data distribution in an end-to-end manner. 
With the popularity of bottom-up attention~\cite{anderson2018bottom}, recent research attentions have been shifted to capturing and modeling relations between entities (\eg, objects or words) via intra-model interaction. As a representative model, VSRN~\cite{li2019visual} reasons across regions in the visual modality by using GCN and Gated Recurrent Unit network for short-long term relation modeling. Qu \etal~\cite{qu2020context} discussed the modality asymmetry issue and proposed a multi-view summarization architecture. Chen \etal~\cite{chen2021learning} discovered that simple global pooling functions can outperform the complicated models and proposed a generalized pooling operator. 
These two-tower models ensure satisfactory retrieval efficiency as features in each modality can be extracted in parallel and indexed in an offline way, but suffer from poor accuracy due to the coarse-grained matching.

In contrast, the one-tower framework~\cite{lee2018stacked, chen2020imram, diao2021similarity, qu2021dynamic} embraces fine-grained cross-modal interactions to achieve more thorough matching between fragments (\eg, objects and words). 
As the pioneering work, SCAN~\cite{lee2018stacked} infers some latent region-word alignments by means of cross-modal interaction. Thereafter, much effort has been dedicated to mining cross-modal matching relations. For instance, Chen \etal~\cite{chen2020imram} incorporated the iteration strategy into SCAN to perform multi-step cross-modal relation reasoning. Liu \etal~\cite{liu2020graph} tailored a graph-based network to explore fine-grained correspondence from node level and structure level. Diao \etal~\cite{diao2021similarity} inferred multi-level semantic correlations in a similarity graph and filtered noisy alignments via the attention mechanism. Differently, Qu \etal~\cite{qu2021dynamic} argued possible optimal patterns may not be explored by existing modality interaction patterns and developed a unified dynamic modality interaction modeling network for automatic routing. 
Despite the superior retrieval accuracy, this line of work has the problem of heavy computational overhead. 

To sum up, most existing works focus on representing data samples in joint embedding space or intensively excavating fine-grained cross-modal alignments, but they ignore the efficient and practical post-processing procedure, \ie,  re-ranking. 

\subsection{Re-ranking}
As an effective and practical technology, re-ranking has attracted a flurry of interest in traditional single-modality retrieval tasks, such as image retrieval~\cite{chum2007total, bai2016sparse, gordo2017end, zhong2017re, zhang2020understanding, ouyang2021contextual, lei2022reducing} and document retrieval~\cite{macavaney2020training, macavaney2020efficient, zhuang2021tilde, zerveas2022mitigating, cai2014personalized, macavaney2019content}. 
A rich line of studies has explored \textbf{Query Expansion} by using the nearest neighbors of a query to generate a new enriched one. For instance, average query expansion (AQE)~\cite{chum2007total} was first proposed to aggregate local features by mean-pooling the top-k images in the original ranking list. Afterward, Gordo~\etal~\cite{gordo2017end} and Radenovic~\etal~\cite{radenovic2018fine} argued that the mean-pooling would induce much noisy information and presented AQEwD and $\alpha$QE, respectively. Another thread of work is \textbf{Diffusion}-based re-ranking strategy, which depends on a neighborhood topology constructed in an offline way and utilizes it at query time for search on the manifold. 
For example, Iscen \etal~\cite{iscen2017efficient}  captured the image manifold in the region-level embedding space and proposed a fast spectral ranking method~\cite{iscen2018fast} to improve the computational efficiency. 
In addition, some research has been devoted to employing neighborhood relations, especially the \textbf{k-Reciprocal Nearest Neighbors}, for re-ranking. For example, Zhong \etal~\cite{zhong2017re} encoded the k-reciprocal neighbors of a query into a single vector and then conducted re-ranking under the Jaccard distance. Liu \etal~\cite{liu2019guided} built a graph based on neighborhood affinities and leveraged GCN for manifold learning. Recently, Ouyang \etal~\cite{ouyang2021contextual} reconstructed affinity features with a set of anchor images for contextual similarity reasoning. As one of the rare efforts to explore re-ranking for image-text retrieval, the k-reciprocal nearest neighbor searching scheme~\cite{wang2019matching} was designed as post-processing to enhance retrieval performance. 
Despite the thrilling progresses, existing methods depend heavily on prior knowledge, lacking sufficient flexibility to adaptively model query-neighbor and neighbor-neighbor relations. Besides, they overlook the sparsity and asymmetry issues widely existing in image-text retrieval. 

\begin{figure*}[t]
	\includegraphics[width=0.93\textwidth]{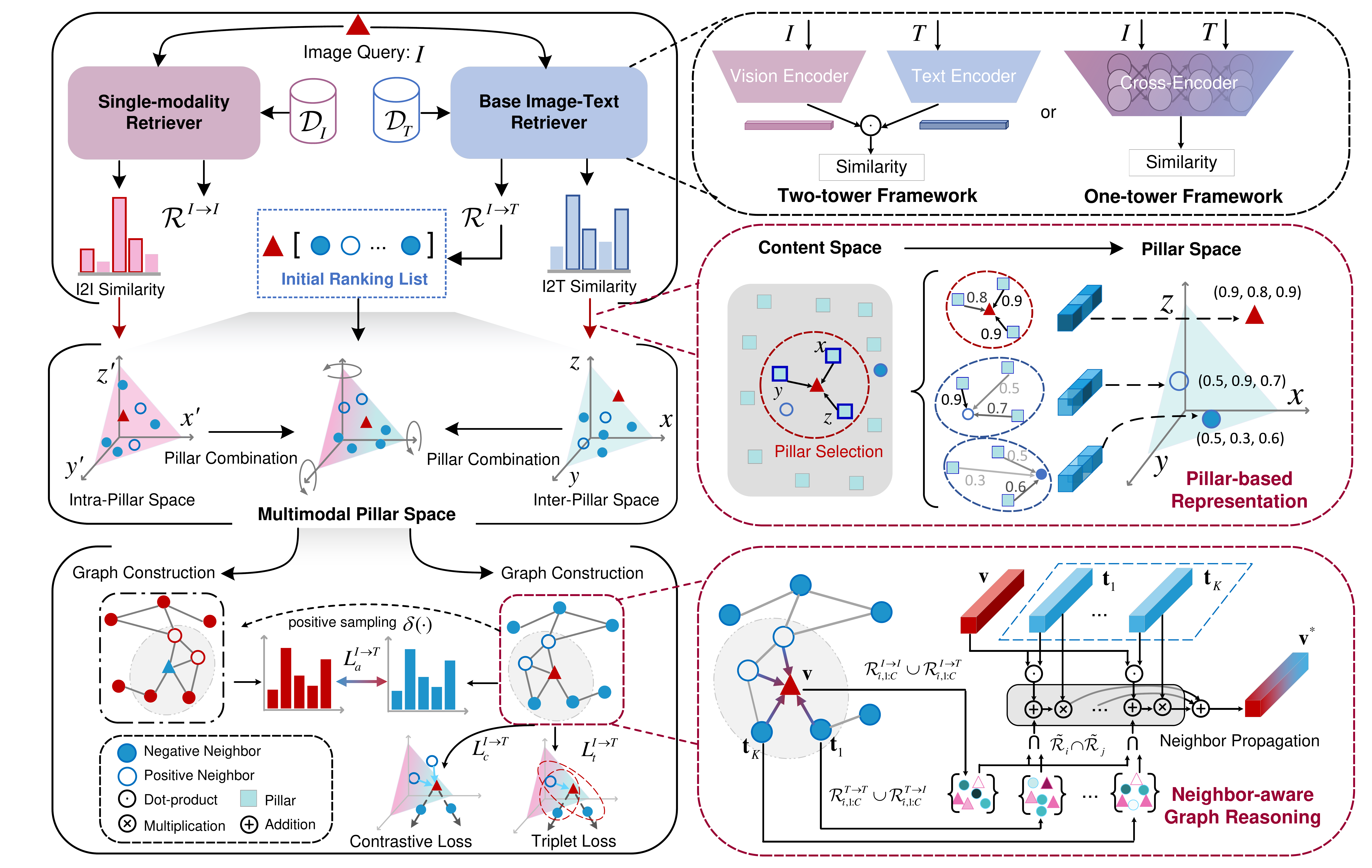}
	\vspace{-2ex}
	\caption{Schematic illustration of the proposed LeaPRR model (image-to-text direction). Given the image query and databases, the initial ranking list is obtained via the base retriever. And then the top-ranked multimodal neighbors are selected as pillars to build the multimodal pillar space. In this space, Pillar-based Representation is conducted for each data sample. Thereafter, the graph is constructed by neighbor-based and learning-based affinities and Neighbor-aware Graph Reasoning is performed. Finally, the model learns neighbor relations for re-ranking by means of three constraints. } 
	\label{fig:pipeline}
	\vspace{-2ex}
\end{figure*}

\section{Methodology}
The overall framework of our proposed LeaPRR is depicted in Figure~\ref{fig:pipeline}. In the following, we will first introduce the problem definition for cross-modal ITR re-ranking and then elaborate on \textit{Pillar-based Representation} and \textit{Neighbor-aware Graph Reasoning}. Finally, we will describe the \textit{Learning} strategy in the pillar space.

\subsection{Problem Definition}
Given a query $q$ (an image or a text) and the database in another modality, \ie, $\mathcal{D}=\{d_1, \ldots, d_C\}$, we first calculate the initial similarity $\phi(q, d_i)$ between the query and each item $d_i$ in $\mathcal{D}$ via a base ITR backbone $\phi$. It can be implemented by a one- or two-tower network. The former~\cite{li2019visual} commonly treats $\phi$ as the cosine similarity between embeddings extracted from two branches, while the latter views $\phi$ as an integrated inter-modal interaction module, such as cross-attention~\cite{chen2020imram, wei2020multi}.
Afterward, we can attain an initial ranking list $\mathcal{R}(q, \mathcal{D})$ by sorting similarities in descending order.

The goal of ITR re-ranking is to refine the initial ranking list so those items that semantically match well with the query are ranked higher and vice versa in the refined ranking list, denoted by $\mathcal{R}^*(q, \mathcal{D})$.
In this work, we focus on re-ranking top-$K$ items in $\mathcal{R}(q, \mathcal{D})$ since users always attach more attention to top-ranked retrieval results in practical scenarios. 

We consider two retrieval settings: Image-to-Text (I2T), \ie, $q := I, \mathcal{D} := \mathcal{D}_T$, and Text-to-Image (T2I), \ie,  $q := T, \mathcal{D} := \mathcal{D}_I$. 
Considering the high consistency except for the retrieval direction, we will elaborate our method under the \textit{I2T setting} without loss of generality. Besides, we will use \textit{items} to denote all retrieved texts (or images) for a given image (or text) query, \textit{neighbors} to refer to top-ranked items, and \textit{entities} to represent any queries or items. 

\subsection{Pillar-based Representation} \label{sec:pillar_rep}

For two-tower I2T frameworks, we can easily access intermediate features of each modality and leverage them to enrich the query or item representation for re-ranking~\cite{radenovic2018fine, gordo2020attention}. Nevertheless, this is not possible with one-tower I2T frameworks due to the existence of cross-modal interaction modules. In the light of this, it is unrealistic to use the intermediate representations to optimize the ranking results of both two-tower and one-tower frameworks. Therefore, we resort to cross-modal similarities for image-text retrieval re-ranking. However, how to represent an image or a text and compare them only depending on similarities, becomes a problem to be solved.

As the saying goes, ``birds of a feather flock together''. We can describe a person with inherent appearance characteristics, such as height, weight, skin color, face shape, and hairstyle.
Meanwhile, we can also depict this person by means of the relation between him and other people.
For instance, it is reasonable to believe that two people are similar in appearance if they have common family members, or similar in interest if with common idols and/or friends.
This inspired us that we can not only represent an entity using its internal content/attributes but also can do it with external relations between it and other entities.

Although others can be employed as references to represent an item, it does not mean anyone can take responsibility. As an illustration, those people that the person is unfamiliar with would not be helpful as references to discriminate him against. Therefore, the relations between an item and the selected references should not be very sparse. 
We will detail how to select such references and exploit them as references to represent an item, in the following. 
\subsubsection{Multimodal Pillar Selection}\label{sec:pillar_selection}

Given the initial ranking list $\mathcal{R}(I, \mathcal{D}_T) = \{T_1, ..., T_N\}$ of I2T retrieval, we select top-$L (L < N)$  texts as the above-mentioned references. In our work, these selected references are defined as \textit{pillars}. 
For the sake of simplicity, we re-denote $\mathcal{R}(I, \mathcal{D}_T)$ as $\mathcal{R}^{I \rightarrow T}$, and utilize $\mathcal{R}_{1:L}^{I \rightarrow T} = \{T_1, ..., T_L\}$ to represent the \textit{inter-modal pillars} of the query $I$.  To further enhance the representation ability, we propose to excavate \textit{intra-modal pillars} from the intra-modal database $\mathcal{D}_I$. Concretely, we apply the image encoder $\psi_{I}$ to extract features and then calculate pairwise similarities between them. After that, based on the intra-modal ranking list $\mathcal{R}(I, \mathcal{D}_I) = \{I_1, ..., I_M\}$, $\mathcal{R}^{I \rightarrow I}$ for short, we collect a set of intra-modal pillars\footnote{In this paper, we use the same number of intra- and inter-modal pillars for the sake of simplicity. In practice, it is acceptable to consider different numbers of pillars.} of query $I$ as $\mathcal{R}_{1:L}^{I \rightarrow I} = \{I_1, ..., I_L\}$. 

\subsubsection{Pillar-based Encoding}
On the basis of pillars, we encode the query image $I$ and its top-$K$ neighbor texts $\{T_1, ..., T_K\}$ as follows:
\begin{equation}
	\begin{split}
		\textbf{v} &= [\oplus_{i=1}^L \phi (I, T_i), \oplus_{j=1}^L \pi (I, I_j)], \\
		\textbf{t}_k & = [\oplus_{i=1}^L \pi (T_k, T_i),  \oplus_{j=1}^L \phi (I_j, T_k)], 
	\end{split}
\end{equation}
where $\oplus$ denotes the concatenation operator, $\textbf{v} \in \mathbb{R}^{2L}$ and $\textbf{t}_k \in \mathbb{R}^{2L}$ respectively represent pillar-based representations of the query $I$ and its $k$-th neighbor $T_k$, $\phi$ can be any base backbone that calculates image-text similarities, and $\pi$ is the intra-modal similarity calculation model. Specifically, $\pi (I, I_j) = \cos (\psi_I (I), \psi_I (I_j) )$, $\pi (T_k, T_i) = \cos (\psi_T (T_k), \psi_T (T_i))$, where $\psi_I$ and $\psi_T$ are image encoder and text encoder, respectively. 

\subsection{Neighbor-aware Graph Reasoning} \label{sec:neighbor_prop}
One of the keys to re-ranking is to model the high-order neighbor relations in a neighborhood including a query and its neighbors. Prior approaches try to achieve it from different aspects. For instance, QE-based models~\cite{radenovic2018fine, arandjelovic2012three} exploit query-neighbor interaction to expand query representation by aggregating neighbors. In contrast, diffusion-based~\cite{iscen2017efficient, pang2018deep} models perform one-way information flow by building a connected graph. Nevertheless, these models are designed based on some rules with limited flexibility. To tackle this issue, we tailor a graph-based neighbor propagation to capture query-neighbor and neighbor-neighbor relations adaptively. 

After finishing pillar encoding, we are able to represent the query and its neighbors in a pillar space. In this space, we view each item (\ie, a query or a neighbor) as a node and then connect them according to their affinity scores.  
\subsubsection{Neighbor-based Affinity}
To exploit neighbor relations from the raw feature space, we first collect top-$C$ neighbors from both intra- and inter-modal ranking lists for each item and then merge them into one set, as,
\begin{equation}
\tilde{\mathcal{R}}_i = 
\begin{cases}
    \mathcal{R}_{i, 1:C}^{I \rightarrow I} \cup \mathcal{R}_{i, 1:C}^{I \rightarrow T}, & \rm node \ \textit{i} \ \rm is \ \rm an \ \rm image, \vspace{1ex}\\
    \mathcal{R}_{i, 1:C}^{T \rightarrow T} \cup \mathcal{R}_{i, 1:C}^{T \rightarrow I}, & \rm node \ \textit{i} \ \rm is \ \rm a \ \rm text. \\
\end{cases}
\end{equation}
where $\tilde{\mathcal{R}}_i$ denotes the neighbor set of the $i$-th node. $\mathcal{R}_{i, 1:C}^{I \rightarrow T}$ refers to the set of top-$C$ neighbors of the $i$-th node in the direction $I \rightarrow T$, and the same is true for other directions. Like $I \rightarrow I$, we also apply a text encoder to obtain $\mathcal{R}_{i, 1:C}^{T \rightarrow T}$.

Afterward, we calculate the edge weight $\textbf{C}_{ij}$ between the $i$-th node and the $j$-th node depending on their common neighbors:
\begin{equation}
    \textbf{\textbf{C}}_{ij} = \frac{|\tilde{\mathcal{R}}_i \cap \tilde{\mathcal{R}}_j|}{\sum_{k=1}^{1+K} |\tilde{\mathcal{R}}_i \cap \tilde{\mathcal{R}}_k|}. 
\end{equation}
where $|\cdot|$ denotes the element number of the set.
To alleviate the influence of noisy edges, we introduce a sparse factor $\lambda$ and the edge weight is updated as follows, 
\begin{equation}\label{eqn:sparse}
\tilde{\textbf{C}}_{ij} = 
\begin{cases}
    \textbf{C}_{ij}, & \textbf{C}_{ij} > \frac{\lambda}{1+K}, \\
    0, & otherwise.
\end{cases}
\end{equation}
Following the row-wise normalization of $\tilde{\textbf{C}}$, we could obtain the neighbor-based affinity matrix, denoted as $\tilde{\textbf{A}} \in \mathbb{R}^{(1+K) \times (1+K)}$.

\subsubsection{Learning-based Affinity}
To endow the model with sufficient flexibility, we drive it to learn informative relations. Specifically, we first stack the pillar-based representations of the query and its neighbors to build a matrix $\textbf{F} = [\textbf{v}; \textbf{t}_1; ...;\textbf{t}_K] \in \mathbb{R}^{(1+K) \times 2L}$. And then we calculate the affinity matrix $\hat{\textbf{A}} \in \mathbb{R}^{(1+K) \times (1+K)}$ as,
\begin{equation}\label{eqn:learning_affinity}
	\hat{\textbf{A}} = {\rm softmax}(f_Q (\textbf{F}) \cdot f_K (\textbf{F})^\top), \\
\end{equation}
where $f_Q$ and $f_K$ respectively denote two fully-connected (FC) layers and ${\rm softmax (\cdot)}$ is performed over each row.  $\hat{\textbf{A}}_{ij}$ represents a learning-based affinity score between node $i$ and node $j$. 

\subsubsection{Neighbor Propagation}
Combining the above two types of affinities, we derive the final affinity matrix by $\textbf{A} = (\tilde{\textbf{A}} + \hat{\textbf{A}}) / 2$ and perform the  neighbor propagation in the first layer of GCN as, 
\begin{equation}\label{eqn:gcn}
    \textbf{F}^{(1)} = g (\textbf{A} \cdot f_V (\textbf{F})) + \textbf{F}, 
\end{equation}
where $f_V$ is an FC layer and $g$ denotes a multi-layer perceptron for feature transformation followed by residual connection. 
After the first layer of GCN, we can 
re-compute the learning-based affinity matrix using Eqn.~(\ref{eqn:learning_affinity}) and then perform the following propagation as Eqn.~(\ref{eqn:gcn}) does. The output of the last layer GCN, denoted as $\textbf{F}^*$, is treated as the refined pillar-based representation matrix. 
\setlength\intextsep{0pt}

\begin{table*}[t]
	\centering
	\setlength{\abovecaptionskip}{0.15cm}
	\caption{Performance comparison between the proposed LeaPRR and several baselines on the Flickr30K and MS-COCO (5K) datasets based on VSE$\infty$ and DIME$^*$. The best performance for each base backbone is highlighted in bold, while the second best results are underlined. The results of the proposed method are marked with a gray background. }
	\label{tab:performance_comparison}
	{\hspace{-1.3ex}
		\resizebox{0.87\textwidth}{!}
		{
			\setlength\tabcolsep{4pt}
			\renewcommand\arraystretch{1.0}
			\begin{tabular}{p{5cm}<{\raggedleft}||ccc|ccc|c||ccc|ccc|c}
				\hline\thickhline
				\multicolumn{1}{c||}{\multirow{3}{*}{Method}} & \multicolumn{7}{c||}{
        \textbf{Flickr30K Dataset}}  & \multicolumn{7}{c}{
        \textbf{MS-COCO (5K) Dataset} } \\ 
				\cline{2-15}  
				\multicolumn{1}{c||}{} & \multicolumn{3}{c|}{Image-to-Text} & \multicolumn{3}{c|}{Text-to-Image} & \multicolumn{1}{c||}{\multirow{2}{*}{rSum}} & \multicolumn{3}{c|}{Image-to-Text} & \multicolumn{3}{c|}{Text-to-Image} & \multicolumn{1}{c}{\multirow{2}{*}{rSum}} \\
				\cline{2-7} \cline{9-14}
				 & R@1 & R@5  & R@10 & R@1 & R@5 & R@10 & & R@1 & R@5  & R@10 & R@1 & R@5 & R@10 & \\
				\hline\hline
				\multicolumn{1}{l||}{$\bullet$~{{VSE}$\infty$}~\pub{CVPR21}~\cite{chen2021learning}}    & 81.7  & 95.4  & 97.6  & 61.4  & 85.9  & 91.5  & 513.5 & 58.3  & 85.3  & 92.3  & 42.4  & 72.7  & 83.2  & 434.2   \\
				\cdashline{1-15}[1pt/1pt]
				{+ AQEwD~\pub{IJCV17}}~\cite{gordo2017end} & 81.7  & 88.8  & 94.0  & 61.4  & 80.0  & 87.0  & 492.9 & 58.3  & 77.5  & 85.1  & 42.4  & 66.3  & 77.5  & 407.1    \\
				{+ $\alpha$QE~\pub{TPAMI18}}~\cite{radenovic2018fine} & 81.7  & 90.9  & 95.7  & 61.4  & 83.4  & 90.7  & 503.8 & 58.3  & 80.6  & 89.3  & 42.4  & 71.0  & 82.1  & 423.7 \\
				\cdashline{1-15}[1pt/1pt]
				+ {$^\dagger$ADBAwd + AQEwD~\pub{IJCV17}}~\cite{gordo2017end} & 79.3  & 88.1  & 92.7  & 37.4  & 69.4  & 79.3  & 446.2  & 55.5  & 75.9  & 83.4  & 29.7  & 58.5  & 71.1  & 374.1   \\
				+ {$^\dagger\alpha$DBA + $\alpha$QE~\pub{TPAMI18}}~\cite{radenovic2018fine} & 80.6  & 90.2  & 95.3  & 59.5  & 79.6  & 88.3  & 493.5  & 59.0  & 79.4  & 87.8  & 41.8  & 68.5  & 80.5  & 417.0  \\
				\cdashline{1-15}[1pt/1pt]
				+ {$^\ddagger$ADBA + AQE~\pub{ICCV07}}~\cite{chum2007total} & 77.0  & 85.2  & 92.4  & 62.7  & 85.6  & 91.2  & 494.1  & 52.0  & 74.1  & 83.4  & 38.5  & 67.5  & 79.1  & 394.6  \\
				+ {$^\ddagger$ADBAwd + AQEwD~\pub{IJCV17}}~\cite{gordo2017end} & 81.7  & 88.9  & 94.0  & \underline{66.4}  & \underline{87.8}  & 92.1  & 510.9  & 58.3  & 77.5  & 85.1  & 42.7  & 70.9  & 81.1  & 415.6  \\
				+ {$^\ddagger\alpha$DBA + $\alpha$QE~\pub{TPAMI18}}~\cite{radenovic2018fine} & 79.0  & 89.4  & 95.4  & 64.8  & \textbf{87.9}  & \textbf{93.3}  & 509.8  & 56.0  & 79.6  & 88.3  & 41.6  & 71.3  & 82.4  & 419.2    \\
				\cdashline{1-15}[1pt/1pt]
				+ {DFS~\pub{CVPR17}}~\cite{iscen2017efficient} & 81.7  & 95.4  & \underline{97.6}  & 61.4  & 86.1  & 91.6  & 513.8   & 58.3  & 85.3  & \underline{92.3}  & 42.4  & 72.7  & 83.2  & 434.3 \\
				+ {FSR~\pub{CVPR18}}~\cite{iscen2018fast} & 80.0  & 94.8  & 97.2  & 60.9  & 83.5  & 87.6  & 504.0  & 56.8  & 84.5  & 92.2  & 41.8  & 71.0  & 80.2  & 426.5  \\
				\cdashline{1-15}[1pt/1pt]
				+ {KRNN~\pub{ACMMM19}}~\cite{wang2019matching} & \underline{84.7}  & \underline{96.4}  & \textbf{98.0}  & 64.6  & 87.7  & \underline{92.2}  & \underline{523.6} & \underline{61.6}  & \underline{87.5}  & \textbf{93.0}  & \underline{45.3}  & \underline{74.5}  & \underline{84.0}  & \underline{445.9}    \\
			\rowcolor{mygray}
                + LeaPRR (Ours) & \textbf{86.2}  & \textbf{96.6}  & \underline{97.6}  & \textbf{66.6}  & \textbf{87.9}  & 91.9  & \textbf{526.7}  &  \textbf{65.0}  & \textbf{87.8}  & \textbf{93.0}  & \textbf{46.8}  & \textbf{74.6}  & \textbf{84.1}  & \textbf{451.5}  \\
				\hline\hline
				\multicolumn{1}{l||}{$\bullet$~{DIME$^*$}~\pub{SIGIR21}~\cite{qu2021dynamic}}   & 81.0    & 95.9  & 98.4  & 63.6  & 88.1  & 93.0    & 520.0  & 59.3  & 85.4  & 91.9  & 43.1  & 73.0    & 83.1  & 435.8 \\
				\cdashline{1-15}[1pt/1pt]
				+ {DFS~\pub{CVPR17}}~\cite{iscen2017efficient} & 80.5  & 93.5  & 95.0  & 59.2  & 72.9  & 75.3  & 476.4  & 59.2  & 85.4  & 91.9  & 42.9  & 70.2  & 78.2  & 427.8  \\
				+ {FSR~\pub{CVPR18}}~\cite{iscen2018fast} & 80.5  & 91.0  & 93.2  & 59.2  & 67.6  & 69.0  & 460.5  & 58.8  & 83.3  & 88.8  & 42.6  & 62.5  & 67.6  & 403.7  \\
				\cdashline{1-15}[1pt/1pt]
				+ {KRNN~\pub{ACMMM19}}~\cite{wang2019matching} & \underline{83.6}  & \textbf{96.8}  & \underline{98.4}  & \underline{65.2}  & \underline{88.3}  & \underline{92.9}  & \underline{525.2}  & \underline{62.9}  & \underline{86.6}  & \underline{92.5}  & \underline{43.8}  & \underline{73.1}  & \underline{83.3}  & \underline{442.2}     \\
                \rowcolor{mygray}
				+ LeaPRR (Ours) & \textbf{86.2}  & \underline{96.7}  & \textbf{98.9}  & \textbf{71.1}  & \textbf{89.5}  & \textbf{93.2}  & \textbf{535.6}  &  \textbf{64.3}  & \textbf{87.3} & \textbf{93.1} & \textbf{45.5} & \textbf{74.6} & \textbf{83.7}  & \textbf{448.5}  \\
				\hline
				\multicolumn{15}{l}{\makecell[l]{$\bullet$: base backbones without re-ranking; $\dagger$: expand database and query sequentially; $\ddagger$: concatenate all queries with database and expand together; \\ $*$: ensemble two single base backbones}} 
			\end{tabular}
		}
	}
    \vspace{-2ex}
\end{table*}
\subsection{Learning from Pillar Space} \label{sec:learning}
Under pairwise supervision between images and texts, our model is optimized with the following three constraints. 
\subsubsection{Global Contrastive Loss}
We adopt the contrastive loss~\cite{hadsell2006dimensionality} to enforce the query close to positives and far away from negatives as much as possible in a global perspective, which is defined as, 
\begin{equation}
	L_c^{I \rightarrow T} = -\log \frac{\sum_{i \in \mathcal{P}} \exp (s(I, T_i) / \tau)}{\sum_{j=1}^{K} \exp(s(I, T_j) / \tau)}, 
\end{equation}
where $\mathcal{P}$ denotes the positive set of $I$, $\tau$ is a scalar temperature factor, and $s(I, T_i) = \cos (\textbf{v}^*, \textbf{t}^*_i)$ refers to the cosine similarity between the refined pillar-based representations of the query image $I$ and its $i$-th textual neighbor $T_i$. 

\subsubsection{Local Triplet Loss}
Different from the contrastive loss that forces similarities of all positive pairs essentially, the hinge-based triplet loss~\cite{schroff2015facenet} focuses on local relative proximity and allows the variance between positives to some extent, making it less greedy. In this paper, considering negative samples in top-$K$ neighbors are already hard, we do not conduct the hardest negative mining but turn to the hardest positive instead, which makes it complementary to the contrastive loss. Formally, we formulate the triplet loss as, 
\begin{equation}
	L_t^{I \rightarrow T} = \sum_{i \in \mathcal{N}} [\alpha - s(I, \hat{T}) + s(I, T_i)]_+, 
\end{equation}
where $\alpha$ is the scalar margin, $\mathcal{N}$ represents the negative set of $I$, and $\mathcal{P} \cup \mathcal{N} = \mathcal{R}^{I \rightarrow T}_{1:K}$. $[x]_+ = \max (x, 0)$ is the hinge function, and $\hat{T}$ is the hardest positive defined as $\hat{T} = \underset{i \in \mathcal{P}}{\arg\min} \ s(I, T_i)$. 

\subsubsection{Mutual Modality Alignment}
Until now, we have illustrated the mechanism of the I2T submodel. Similarly, we can derive the T2I submodel by respectively assigning the query and database as the text $T$ and $\mathcal{D}_I$.

To ensemble these two submodels, a naive way is to directly average two similarities, as most of prior base backbones~\cite{lee2018stacked, qu2021dynamic, liu2020graph} do. In this way, however, two submodels are trained independently and unable to promote each other. Inspired by mutual learning~\cite{zhang2018deep, wen2021comprehensive}, we devise a mutual modality alignment scheme to spur two submodels to teach and promote each other together. 

Taking I2T as an example, we first carry out cross-modal positive sampling for the query $I$ and each neighbor $T_i$ in $\mathcal{D}_T$, obtaining  $T' = \delta(I)$ and $I'_i = \delta(T_i)$, where $\delta(\cdot)$ is the sampling function. Concretely, for a given entity, we first find its positive matches in another modality by the pairwise ground-truth information and then uniformly sample one if there are multiple positives. Afterward, we respectively calculate two similarity distributions centered with $I$ and $T'$, as, 
\begin{equation}
	\begin{split}
		\textbf{p}_I &= \rm softmax (s(I, T_1)/\tau, ..., s(I, T_K)/\tau), \\
		\textbf{q}_{T'} &= \rm softmax (s(T', I'_1)/\tau, ..., s(T', I'_K)/\tau). 
	\end{split}
\end{equation}
Fianlly, the alignment loss $L_a$ is defined based on Kullback–Leibler divergence $D_{KL}$ as, 
\begin{equation}
	L_a^{I \rightarrow T} = D_{KL} (\textbf{p}_I || \textbf{q}_{T'}).
\end{equation}

By combining all the above three loss functions in I2T and T2I settings, we could acquire the total loss for model optimization, as, 
\begin{equation}
	L = \sum_{j \in \{I \rightarrow T, T \rightarrow I\}} \sum_{i \in \{c, t, a\}} L_i^j.
\end{equation}

\section{Experiments}
In this section, we conducted experiments on two benchmark datasets to answer the following research questions:
\begin{itemize}[leftmargin=*]
    \item \textbf{RQ1}: How does LeaPRR perform in the image-text retrieval task compared with state-of-the-art baselines?
    \item \textbf{RQ2}: How does each component of LeaPRR affect the retrieval performance?
    \item \textbf{RQ3}: How are the generalization and transferability of LeaPRR when facing different base ITR backbones and datasets? 
\end{itemize}

\subsection{Datasets}\label{sec:dataset}
We validate our method on two large-scale datasets: Flickr30K~\cite{young2014image} and MS-COCO~\cite{lin2014microsoft}.

\textbf{Flickr30K}~\cite{young2014image}: It comprises 31,783 images collected from Flickr, and each image corresponds to 5 human-annotated sentences. We employ the same split as \cite{li2019visual, qu2020context} did, specifically, 29,783 training images, 1,000 validation images, and 1,000 testing images. 

\textbf{MS-COCO}~\cite{lin2014microsoft}: It includes 123,287 images and each is annotated with 5 sentences. Following the split of \cite{lee2018stacked, li2019visual, qu2021dynamic}, we use 113,287 images for training, 5,000 images for validation, and 5,000 images for testing. In particular, we adopt the challenging evaluation setting \textbf{MS-COCO (5K)}, \ie, directly testing on the full 5K images.

\subsection{Experimental Settings}
\subsubsection{Evaluation Protocols}
Following the existing work \cite{li2019visual, chen2021learning, qu2021dynamic}, we conduct evaluation by the standard recall metric R@K (K=1, 5, and 10) and rSum. Concretely, R@K refers to the percentage of queries for which the corrected item is retrieved within the top-K of the ranking list and rSum means the sum of all R@K in both image-to-text and text-to-image retrieval directions for overall performance evaluation. The higher R@K and rSum are better.

\subsubsection{Implementation Details}
In our work, 64 intra-modal pillars and 64 inter-modal pillars are used to construct a 128-dimensional pillar space, \ie, $L=64$. Top-32 and Top-8 items in I2T and T2I settings are considered for re-ranking, respectively. In other words, $K=32$ for I2T and $K=8$ for T2I. Moreover, the dimension of intermediate embedding space mapped via FC layers is set to 768. The sparse factor in Eqn.~(\ref{eqn:sparse}) is set as 0.8. The number of GCN layers is set to 2 for neighbor propagation. The margin $\alpha$ in local triplet loss and the temperature factor $\tau$ are set as 0.2 and 1.0, respectively. In addition, we used the SGD optimizer with a momentum of 0.9 and batch size of 512. Our model is trained for 30 epochs with a learning rate of 0.01. The checkpoint with the highest rSum on the validation set is chosen for testing. 

In this paper, we directly applied the visual and textual branches of the pre-trained two-tower model CAMERA\cite{qu2020context}
published in the open-source community\footnote{\url{https://acmmmcamera.wixsite.com/camera}.} to calculate intra-modal similarities and collected intra-modal pillars, \ie, $\mathcal{R}_{1:L}^{I \rightarrow I}$ and $\mathcal{R}_{1:L}^{T \rightarrow T}$ for all base backbones and experiments. 

\subsection{Performance Comparison (RQ1)}
This section demonstrates the comparison with other state-of-the-art re-ranking methods, including query expansion based methods (\textbf{AQE}~\cite{chum2007total}, \textbf{AQEwD}~\cite{gordo2017end}, \textbf{$\alpha$QE}~\cite{radenovic2018fine}, and them combined with database-side augmentation (\textbf{DBA})~\cite{arandjelovic2012three}\footnote{\url{https://github.com/naver/deep-image-retrieval}.}, including average augmentation (\textbf{ADBA}), with-decay augmentation (\textbf{ADBAwD}), and weighted augmentation (\textbf{$\alpha$DBA}), respectively), diffusion based methods (\textbf{DFS}~\cite{iscen2017efficient}, \textbf{FSR}~\cite{iscen2018fast})\footnote{\url{https://github.com/ducha-aiki/manifold-diffusion}.}, and reciprocal nearest neighbor based ones (\textbf{KRNN}~\cite{wang2019matching})\footnote{\url{https://github.com/Wangt-CN/MTFN-RR-PyTorch-Code}.}. These re-ranking baselines are implemented on top of two representative state-of-the-art base backbones, including a two-tower base backbones \textbf{VSE$\infty$}\footnote{\url{https://github.com/woodfrog/vse_infty}.}~\cite{chen2021learning} and a one-tower base backbone \textbf{DIME$^*$}\footnote{\url{https://sigir21.wixsite.com/dime}.}~\cite{qu2021dynamic}. Note that we directly utilized the published pre-trained parameters to calculate image-text similarities to train and evaluate LeaPRR and other re-ranking baselines. Besides, we inherit ensemble similarities (\ie, averaging two similarities calculated by two single base backbones) for challenging evaluation if available. As query expansion re-ranking baselines (\eg, AQE, AQEwD, $\alpha$QE, and their corresponding DBA versions) require accessing intermediate intra-modal representations, we can only applied them to two-tower base backbones. 
 
\begin{table}[t]
	\centering
	\setlength{\abovecaptionskip}{0.1cm}
	\caption{Ablation Study on Flickr30K regarding different pillar selection strategies. Pil. denotes Pillar.}
	\label{tab:ablation_pil}
	{\hspace{-1.3ex}
		\resizebox{0.35\textwidth}{!}
		{
			\setlength\tabcolsep{3.4pt}
			\renewcommand\arraystretch{1.0}
				\begin{tabular}{l||cc|cc}
					\hline\thickhline 
					& \multicolumn{2}{c|}{Image-to-Text} & \multicolumn{2}{c}{Text-to-Image} \\
					\multicolumn{1}{c||}{\multirow{-2}{*}{Method}} &  R@1 & R@5 & R@1 & R@5 \\
					\hline\hline
					\multicolumn{1}{l||}{{DIME$^*$}~\pub{SIGIR21}~\cite{qu2021dynamic}} & 81.0    & 95.9  & 63.6  & 88.1 \\
					\ \ + LeaPRR (Bottom Pil.) & 78.8  & 94.4  & 64.8  & 87.3 \\
					\ \ + LeaPRR (Random Pil.) & 81.8  & 95.3  & 62.9  & 87.3 \\
					\ \ + LeaPRR (w/o Inter-Pil.) & 80.3  & 96.2  & 66.7  & 89.0 \\
					\ \ + LeaPRR (w/o Intra-Pil.) & 84.3  & 95.9  & 67.9  & 89.1 \\
					\ \ + LeaPRR & \textbf{86.2}  & \textbf{96.7}  & \textbf{71.1}  & \textbf{89.5} \\
					\hline
				\end{tabular}
			}
		}
    \vspace{-2ex}
\end{table}

The comparisons are summarized in Table~\ref{tab:performance_comparison}. From this table, we have the following observations:
\begin{itemize}[leftmargin=*]
    \item Directly applying Query Expansion methods~\cite{gordo2017end, radenovic2018fine} to the two-tower base backbone, \ie, VSE$\infty$, is not helpful. Concretely, after executing QE-based re-ranking for VSE$\infty$, the performance drops, especially on R@5 and R@10. The main reasons are as follows: 1) The intrinsic sparsity of ITR (discussed in Section~\ref{sec:introduction}) may induce top-ranked negative neighbor information to be integrated into the query, therefore overwhelming the informative representation of the original query;
    and 2) the modality asymmetry makes it difficult to fuse neighbors from a different modality. 
    Besides, performing DBA before QE (\eg, $^\dagger\alpha$DBA + $\alpha$QE) fails to redeem this but even makes it worsen, which is mainly attributable to the sparsity and asymmetry issues that have not been tackled. 
    \item By combining DBA and QE at the same time for re-ranking, the performance on some metrics (\eg, recall values at T2I on Flickr30K) gets improved. Among them, $^\ddagger$ADBAwd + AQEwD~\cite{gordo2017end} gains the better R@1 (5.0\% higher than VSE$\infty$)  and $^\ddagger\alpha$DBA + $\alpha$QE~\cite{radenovic2018fine} performs better on R@5 (2.0\% gain) and R@10 (1.8\% gain). The reason may be that the combination strategy relieves the asymmetry problem. In other words, the query could absorb certain informative clues from intra-modal neighbors.
    \item The diffusion approach DFS~\cite{iscen2017efficient} gains marginal overall performance improvement (0.3\% and 0.1\% regarding rSum on Flickr30K and MS-COCO (5K), respectively) compared with the base VSE$\infty$~\cite{chen2021learning}. However, it fails to offer consistent improvement for DIME$^*$~\cite{qu2021dynamic} due to the possible more complicated manifold structure caused by cross-modal interactions in this one-tower architecture. 
    \item The reciprocal neighbor-based method KRNN~\cite{wang2019matching}, acting as a strong baseline, surpasses all aforementioned baselines in terms of most metrics. This demonstrates the effectiveness of capturing and modeling reciprocal query-neighbor relations for ITR. 
    \item LeaPRR outperforms all the baselines over most metrics on top of two base backbones, \eg, it gains 15.9\% and 12.7\% improvement regarding rSum for DIME$^*$~\cite{qu2021dynamic} on Flickr30K and MS-COCO (5K), respectively. The remarkable and consistent improvement is attributed to 1) the excellent \textbf{flexibility} endowed by pillar-based representation and neighbor-aware graph reasoning, and 2) the settlement of the \textbf{sparsity} and \textbf{asymmetry} challenges powered by multimodal pillars, local-global constraints, and the mutual modality alignment. 
\end{itemize}

\begin{table}[t]
    \centering
    \setlength{\abovecaptionskip}{0.1cm}
    \caption{Ablation Study on Flickr30K regarding different constraints and ensemble ways. Cont., Trip., and MMA stand for global contrastive loss, local triplet loss, and mutual modality alignment, respectively.}
    \label{tab:ablation_loss}
    {\hspace{-1.3ex}
        \resizebox{0.37\textwidth}{!}
        {
            \setlength\tabcolsep{3.4pt}
            \renewcommand\arraystretch{1.0}
                \begin{tabular}{l||cc|cc}
                    \hline\thickhline 
                    & \multicolumn{2}{c|}{Image-to-Text} & \multicolumn{2}{c}{Text-to-Image} \\
                    \multicolumn{1}{c||}{\multirow{-2}{*}{Method}} &  R@1 & R@5 & R@1 & R@5 \\
                    \hline\hline
                    \multicolumn{1}{l||}{{DIME$^*$}~\pub{SIGIR21}~\cite{qu2021dynamic}} & 81.0    & 95.9  & 63.6  & 88.1 \\
                    \ \ + LeaPRR (w/o Cont. Loss) & 86.1  & 95.9  & 70.4  & 89.4 \\
                    \ \ + LeaPRR (w/o Trip. Loss) & 85.5  & 96.1  & 69.9  & 88.7 \\
                    \ \ + LeaPRR (w/o MMA, only i-t) & 83.0    & 95.0    & 70.1  & 89.3 \\
                    \ \ + LeaPRR (w/o MMA, only t-i) & \textbf{86.7}  & \textbf{97.8}  & 66.9  & 89.2 \\
                    \ \ + LeaPRR (w/o MMA, i-t \& t-i) & 86.0    & 96.8  & 68.7  & \textbf{89.5} \\
                    \ \ + LeaPRR & 86.2  & 96.7  & \textbf{71.1}  & \textbf{89.5} \\
                    \hline
                \end{tabular}
            }
        }
        \vspace{-5ex}
\end{table}

\subsection{In-depth Analysis (RQ2 \& RQ3)} 

\subsubsection{Ablation Study}
On top of the challenging state-of-the-art base backbone DIME$^*$, We conduct a series of experiments on Flickr30K to verify the effectiveness of each component of LeaPRR and explore their impacts as follows.

\noindent$\bullet$~\textbf{Impact of Pillar-based Represen.}
Considering our proposed model performs a series of learning in the pillar space, the quality of this space plays a key role in the overall performance. In this section, we hence first  justified the effectiveness of our multimodal pillar selection strategy by designing the following variants:
1) \textbf{Bottom Pil.}, collecting $L$ items located in the tail of the intra- and inter-modal ranking list 
to constitute the pillar set; 2) \textbf{Random Pil.}, randomly and uniformly sampling $L$ items from the intra- and inter-model databases; 3) \textbf{w/o Inter-Pil.}, removing inter-modal pillars;
and 4) \textbf{w/o Intra-Pil.}, excluding intra-modal pillars.

From Table~\ref{tab:ablation_pil}, we observe that the first two variants Bottom Pil. and Random Pil. 
get performance degradation as compared to the full model, indicating that inferior pillars would deteriorate retrieval performance.  
Nevertheless, the performance achieved by these two variants is still comparable with the base backbone DIME$^*$~\cite{qu2021dynamic}, demonstrating that our LeaPRR can tolerate bad pillars and avoid performance collapse. 
Besides, the improvements achieved by the full model LeaPRR compared with  two single-pillar variants w/o Intra-Pil and w/o Inter-Pil. verifies the effectiveness of the top-$L$ pillars selected from the intra- or inter-modal database. 
The remarkable performance improvement achieved by DIME$^*$+LeaPRR strongly attests to the effectiveness of our proposed selection strategy and the complementary intra-modal and inter-modal pillars. 

\begin{figure*}[t]
	\subfigure[Number of Pillars.]{
		\label{fig:num_pillars}
		\includegraphics[width=0.305\textwidth]{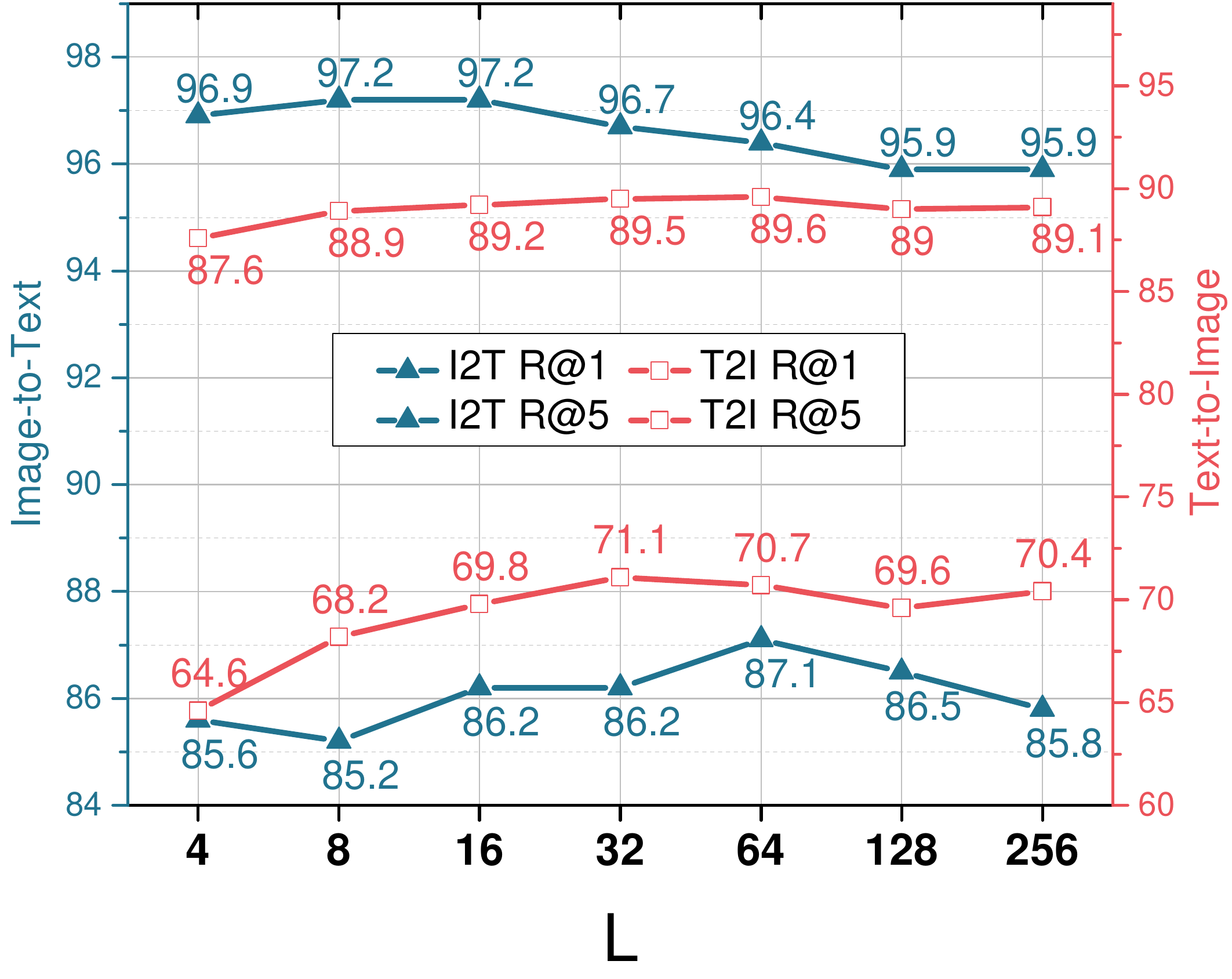}}	
	\subfigure[Number of Neighbors.]{
		\label{fig:num_neighbors}
		\includegraphics[width=0.28 \textwidth]{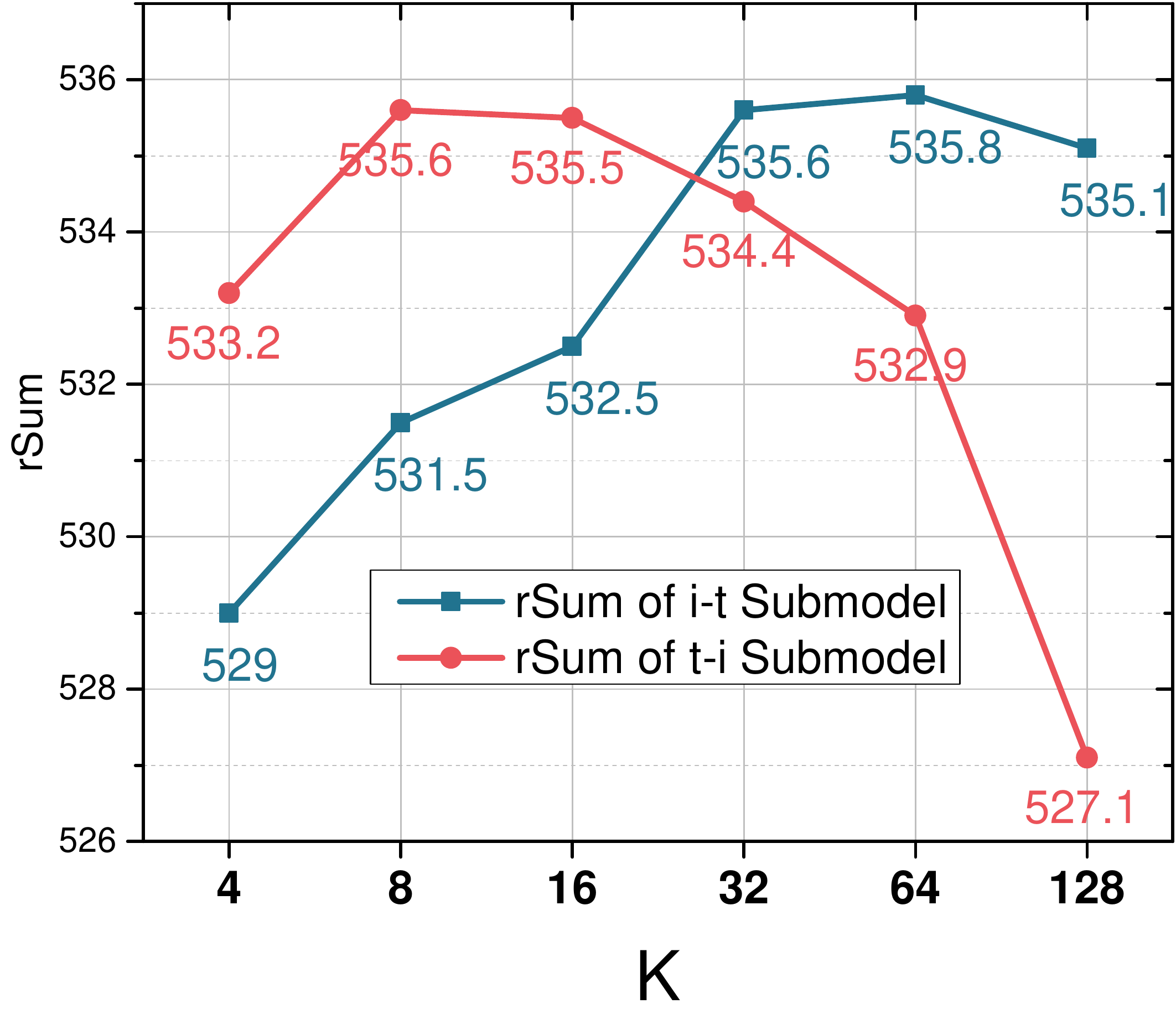}}
        \subfigure[Neighbor-aware Graph Reasoning.]{
		\label{fig:abla_gcn}
		\includegraphics[width=0.31 \textwidth]{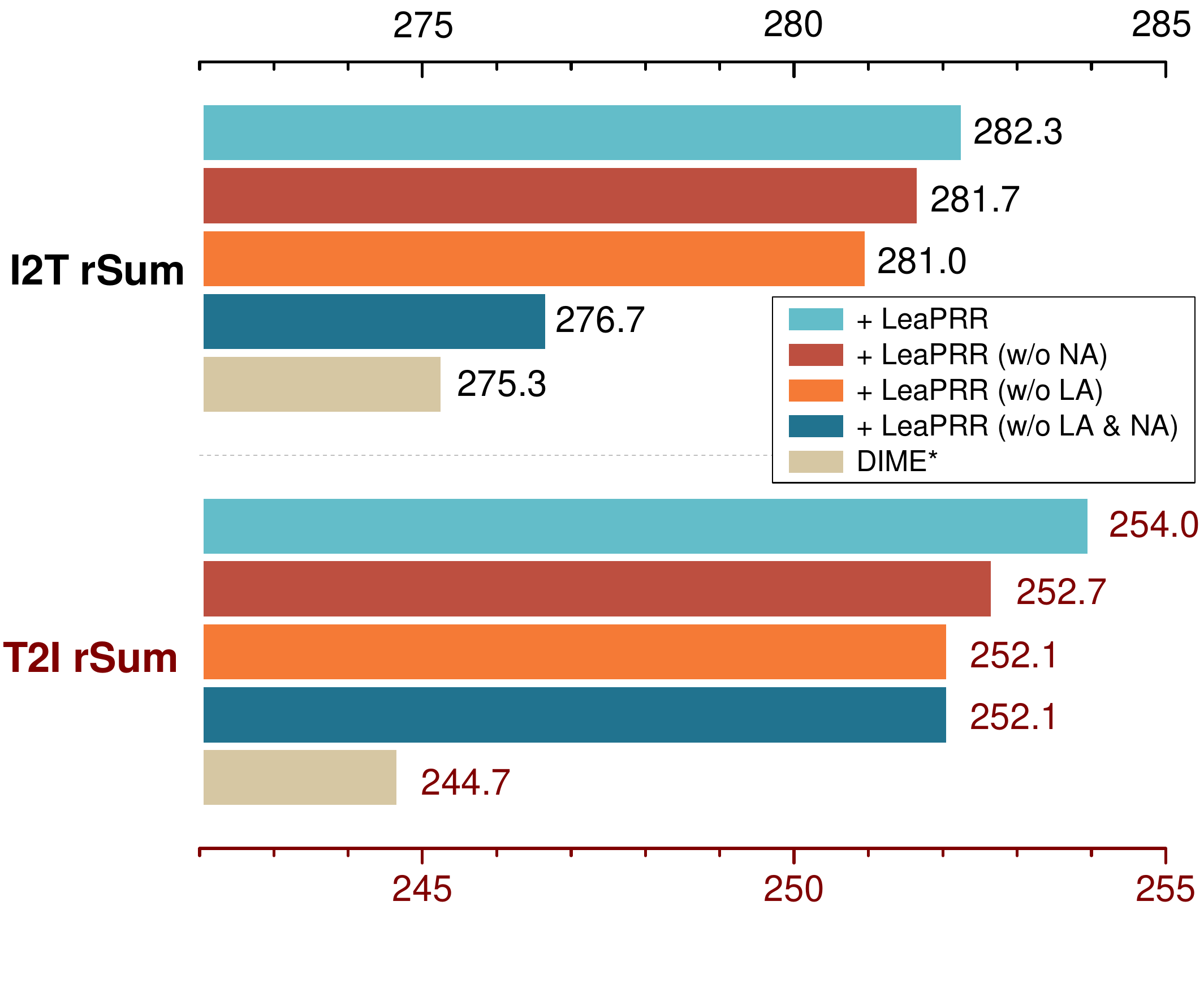}}
	\vspace{-3.0ex}
	\caption{Impact of (a) the pillar number $L$, (b) the neighbor number $K$, and (c) the neighbor-aware graph reasoning module on retrieval performance. Evaluation performed on the Flickr30K dataset and the base backbone DIME$^*$~\cite{qu2021dynamic}. NA and LA in (c) denote neighbor-based affinity and learning-based affinity, respectively. }
	\label{fig:pillars_neighbors_gcn}
\end{figure*}

\noindent$\bullet$~\textbf{Impact of Neighbor-aware Graph Reasoning.}
To assess how the neighbor-aware graph module in Section~\ref{sec:neighbor_prop} contributes to the performance, we design three variants to compare with the base backbone DIME$^*$ and LeaPRR by removing the neighbor-based affinity, the learning-based affinity, and both. As displayed in Figure~\ref{fig:abla_gcn}, the full model achieves the best I2T and T2I retrieval performance compared with other variants, which demonstrates the effectiveness of the neighbor-aware graph reasoning module. Besides, we can observe \textbf{w/o NA} performs better than \textbf{w/o LA}, showing learning to estimate edge weights and construct the graph in the pillar space is more important, because it endows LeaPRR with flexibility. Finally, by comparing the two retrieval directions, more performance improvement by contrasting the full model and \textbf{w/o LA \& NA} is achieved in I2T. One of the reasons is that the positive distribution is sparser in T2I than I2T. From this perspective, it verifies the existence of the sparsity challenge in Section~\ref{sec:introduction}.

\noindent$\bullet$~\textbf{Impact of Learning Objectives.}
To thoroughly analyze the effectiveness of the three learning objectives presented in Section~\ref{sec:learning}, we carried out ablation experiments by eliminating each of them. Concretely, we separately remove the global contrastive objective (\textbf{w/o Cont. Loss}), the local triplet objective (\textbf{w/o Trip. Loss}), and mutual modality alignment (\textbf{w/o MMA}). Besides, as to MMA, we further explored the contribution of each branch by building the following variants: 1) \textbf{w/o MMA, only i-t}, 2) \textbf{w/o MMA, only t-i}, and 3) \textbf{w/o MMA, i-t \& t-i} 
(\ie, separately training two single models and then combining them).

By analyzing the comparison results reported in Table~\ref{tab:ablation_loss}, we have the following observations: 1) The contrastive objective and the triplet objective are able to enhance the learning of LeaPRR from different perspectives and collaborate with each other to improve the performance. Considering most of the raw content information has been abandoned in the pillar space, sufficient optimization driven by these two constraints is crucial; 2) Compared with the contrastive one, the triplet objective seems to contribute more, which reflects the importance of focusing on local hard structures (\eg, relations among the query, hard positives, and hard negatives); 3) Two single submodels still gain obvious performance improvement in terms of all recall metrics strongly validate the effectiveness of our proposed re-ranking framework; 4) The MMA module plays a critical role in combining two single submodels, aligning cross-modal local structures, and mutual enhancement; And 5) considerable overall performance (rSum) improvement can be gained, though some of metrics (\eg, R@5 in I2T) may sacrifice in ensemble models.  

\subsubsection{Parameter Sensitivity} 
To further delve into how LeaPRR performs with regard to pillars and neighbors, we conducted extensive experiments using varying numbers of pillars and neighbors.

\noindent$\bullet$~\textbf{Impact of the Pillar Number ($L$).}
As shown in Figure~\ref{fig:num_pillars}, we evaluated and compared the retrieval performance by selecting various numbers of top-ranked pillars with a fixed exponential interval. From the compared results, we found that saturation points exist in all metrics, roughly at $L=32$ or $64$, except for I2T R@5. 
Too few pillars are inadequate to represent data samples, while too many pillars have a higher risk to introduce noisy pillars.
From this knowable, an appropriate number of pillars is beneficial to make pillar representations rich and discriminative. More importantly, almost all the settings improve the base backbone from different depths, further indicating the effectiveness and robustness. 

\noindent$\bullet$~\textbf{Impact of the Neighbor Size ($K$).}
To investigate the sensitivity of LeaPRR to neighbors, we evaluated its performance under different numbers of neighbors, \ie, $K$, in each submodel and reported the corresponding results in Figure~\ref{fig:num_neighbors}.

These results show that: 1) The overall performance, \ie, rSum, improves with the increasing neighbors at the beginning. When the number increases to a range of values for each submodel, the performance reaches the saturation point and then decreases. This indicates that exploiting a proper number of neighbors can enhance performance and increase the probability that bottom-ranked positives are recalled.
And 2) compared with the i-t submodel, the t-i one gets saturated and starts to rapidly deteriorate earlier. It may attribute to the intrinsic ratio of positives to negatives in existing datasets. As described in Section~\ref{sec:dataset}, each image in both datasets corresponds to five sentences, \ie more negative samples exist in the graph and the top-K neighbor set for the t-i submodel than for the i-t one. 
More importantly, regardless of the number of neighbors, all the settings manage to achieve overall performance improvement over the base backbone, which also verifies the effectiveness and robustness of LeaPRR. 

\begin{table}[t]
	\centering
	\setlength{\abovecaptionskip}{0.1cm}
	\caption{Quantitative results on the Flickr30K and MS-COCO (5K) datasets based on five base ITR architectures for generalization evaluation. The performance improvement compared with base backbones achieved by the proposed LeaPRR is marked in {\color{mygreen}green}.}
	\label{tab:generalization}
	{\hspace{-1.3ex}
		\resizebox{0.37\textwidth}{!}
		{
			\setlength\tabcolsep{3.3pt}
			\renewcommand\arraystretch{0.9}
			\begin{tabular}{l||cc|cc}
				\hline\thickhline 
				& \multicolumn{2}{c|}{Image-to-Text} & \multicolumn{2}{c}{Text-to-Image} \\
				\multicolumn{1}{c||}{\multirow{-2}{*}{Method}} &  R@1 & R@5 & R@1 & R@5 \\
				\hline\hline
                    ~ & \multicolumn{4}{c}{(Flickr30K)} \\
				\multicolumn{1}{l||}{{SCAN$^*$}~\pub{ECCV18}~\cite{lee2018stacked}}   & 69.0  & 89.9  & 47.8  & 77.7     \\
				\small{ \ \ + {KRNN~\pub{ACMMM19}}~\cite{wang2019matching}} & 69.6  & 92.4  & 52.0  & 79.3     \\
				\small{ \ \ + LeaPRR} (Ours) & \reshl{75.1}{6.1}  & \reshl{93.4}{3.5}  & \reshl{57.8}{10.0}  & \reshl{79.3}{1.6}    \\
				\hline
				\multicolumn{1}{l||}{{VSRN$^*$}~\pub{ICCV19}~\cite{li2019visual}}   & 71.5  & 90.8  & 54.8  & 80.9     \\
				\small{ \ \ + {KRNN~\pub{ACMMM19}}~\cite{wang2019matching}} & 76.0  & 94.1  & 55.9  & 81.6   \\
				\small{ \ \ + LeaPRR} (Ours) & \reshl{78.8}{7.3}  & \reshl{93.6}{2.8}  & \reshl{61.0}{6.2}  & \reshl{82.5}{1.6}    \\
				\hline
				\multicolumn{1}{l||}{{CAMERA$^*$}~\pub{ACMMM20}~\cite{qu2020context}}   & 78.0  & 95.1  & 60.3  & 85.9     \\
				\small{ \ \ + {KRNN~\pub{ACMMM19}}~\cite{wang2019matching}} & 81.5  & 96.4  & 63.2  & 87.5    \\
				\small{ \ \ + LeaPRR} (Ours) & \reshl{84.9}{6.9}  & \reshl{96.8}{1.7}  & \reshl{64.6}{4.3}  & \reshl{87.4}{1.5}   \\
				\hline
				\multicolumn{1}{l||}{{{VSE}$\infty$}~\pub{CVPR21}~\cite{chen2021learning}}   & 81.7  & 95.4  & 61.4  & 85.9     \\
				\small{ \ \ + {KRNN~\pub{ACMMM19}}~\cite{wang2019matching}} & 84.7  & 96.4  & 64.6  & 87.7    \\
				\small{ \ \ + LeaPRR} (Ours) & \reshl{86.2}{4.5}  & \reshl{96.6}{1.2}  & \reshl{66.6}{5.2}  & \reshl{87.9}{2.0}    \\
				\hline
				\multicolumn{1}{l||}{{DIME$^*$}~\pub{SIGIR21}~\cite{qu2021dynamic}}   & 81.0    & 95.9   & 63.6  & 88.1   \\
				\small{ \ \ + {KRNN~\pub{ACMMM19}}~\cite{wang2019matching}} & 83.6  & 96.8  & 65.2  & 88.3  \\
				\small{ \ \ + LeaPRR} (Ours) & \reshl{86.2}{5.2}  & \reshl{96.7}{0.8}  & \reshl{71.1}{7.5}  & \reshl{89.5}{1.4}    \\
				\hline \hline
                    ~ & \multicolumn{4}{c}{(MS-COCO (5K))} \\
                    \multicolumn{1}{l||}{{SCAN$^*$}~\pub{ECCV18}~\cite{lee2018stacked}}   & 47.3  & 78.2  & 35.3  & 65.9    \\
                    \small{ \ \ + {KRNN~\pub{ACMMM19}}~\cite{wang2019matching}} & 53.4  & 82.1  & 38.2  & 68.4    \\
                    \small{ \ \ + LeaPRR} (Ours) & \reshl{58.1}{10.8}  & \reshl{84.2}{6.0}  & \reshl{41.3}{6.0}  & \reshl{69.4}{3.5}    \\
                    \hline
                    \multicolumn{1}{l||}{{VSRN$^*$}~\pub{ICCV19}~\cite{li2019visual}}   & 53.0  & 81.1  & 40.5  & 70.6     \\
                    \small{ \ \ + {KRNN~\pub{ACMMM19}}~\cite{wang2019matching}} & 56.3  & 84.1  & 41.1  & 71.0    \\
                    \small{ \ \ + LeaPRR} (Ours) & \reshl{60.0}{7.0}  & \reshl{84.8}{3.7}  & \reshl{43.0}{2.5}  & \reshl{71.7}{1.1}    \\
                    \hline
                    \multicolumn{1}{l||}{{CAMERA$^*$}~\pub{ACMMM20}~\cite{qu2020context}}   & 55.1  & 82.9  & 40.5  & 71.7     \\
                    \small{ \ \ + {KRNN~\pub{ACMMM19}}~\cite{wang2019matching}} & 57.9  & 85.2  & 43.0  & 73.6  \\
                    \small{ \ \ + LeaPRR} (Ours) & \reshl{61.5}{6.4}  & \reshl{85.9}{3.0}  & \reshl{44.1}{3.6}  & \reshl{73.0}{1.3}    \\
                    \hline
                    \multicolumn{1}{l||}{{{VSE}$\infty$}~\pub{CVPR21}~\cite{chen2021learning}}   & 58.3  & 85.3  & 42.4  & 72.7     \\
                    \small{ \ \ + {KRNN~\pub{ACMMM19}}~\cite{wang2019matching}} & 61.6  & 87.5  & 45.3  & 74.5    \\
                    \small{ \ \ + LeaPRR} (Ours) & \reshl{65.0}{6.7}  & \reshl{87.8}{2.5}  & \reshl{46.8}{4.4}  & \reshl{74.6}{1.9}  \\
                    \hline
                    \multicolumn{1}{l||}{{DIME$^*$}~\pub{SIGIR21}~\cite{qu2021dynamic}}   & 59.3  & 85.4  & 43.1  & 73.0     \\
                    \small{ \ \ + {KRNN~\pub{ACMMM19}}~\cite{wang2019matching}} & 62.9  & 86.6  & 43.8  & 73.1    \\
                    \small{ \ \ + LeaPRR} (Ours) & \reshl{64.3}{5.0}  & \reshl{87.3}{1.9}  & \reshl{45.5}{2.4}  & \reshl{74.6}{1.6}    \\
                    \hline
				\multicolumn{5}{l}{\makecell[l]{$*$: ensemble two single base backbones}} 
			\end{tabular}
		}
	}
    \vspace{-3.0ex}
\end{table}

\subsubsection{Generalization Analysis} 
As discussed in Section~\ref{sec:introduction}, generalization is essential for re-ranking. To verify whether LeaPRR can perform well for different base ITR architectures, we plug it into five representative ones: two one-tower (SCAN$^*$ and DIME$^*$) and three two-tower (VSRN$^*$, CAMERA$^*$, and VSE$\infty$) architectures and compare it with KRNN~\cite{wang2019matching}. Results on Flickr30K and MS-COCO (5K) are reported in Table~\ref{tab:generalization}. The superior performance, especially the R@1 performance, on both datasets validates the strong generalization capability of our method and the superiority over the current state-of-the-art KRNN, regardless of base backbones.

\begin{figure}[t]
        \vspace{-2ex}
	\includegraphics[width=0.35\textwidth]{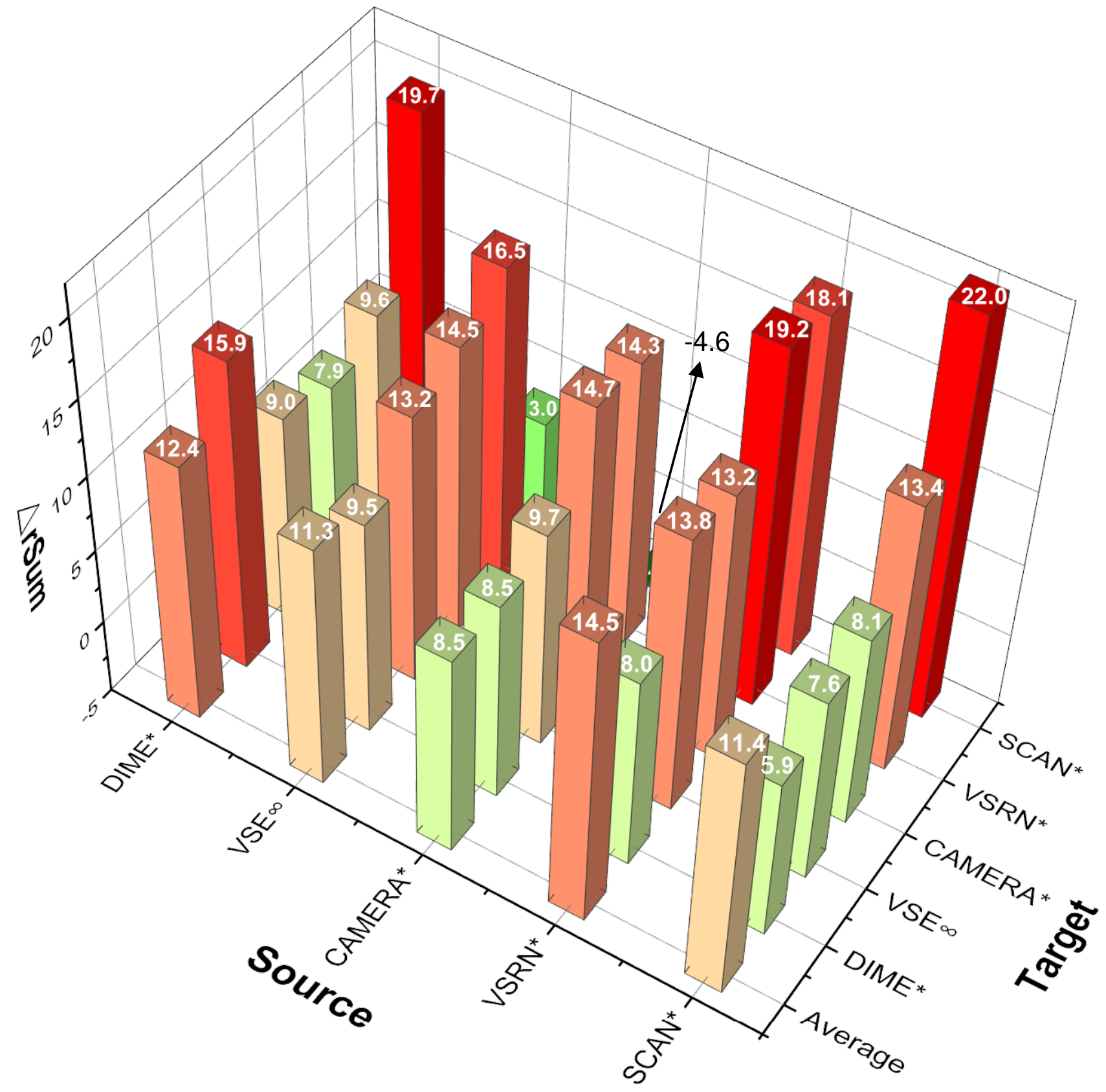}
	\vspace{-2ex}
	\caption{Results on Flickr30K based on five base ITR models for transferability evaluation. LeaPRR are trained on similarities calculated by source base backbones in the training set, and then are evaluated on similarities of the test set calculated by target base backbones. $\Delta$ rSum means the improvement achieved by LeaPRR compared with corresponding base backbones after being transferred. }
	\label{fig:model_transferability}
	\vspace{-4ex}
\end{figure}

\subsubsection{Transferability Analysis} 
The transferability includes backbone transferability and dataset transferability. These two reflect the zero-shot ability across different backbones and datasets, respectively. 
The bulk of previous efforts on ITR focuses on representing and aligning visual and textual content. Due to the existence of a large discrepancy between content spaces, these methods suffer from limited transferability. In contrast, our proposed LeaPRR goes beyond raw content information and embraces pillars to exploit relations between entities for representation. We believe such pillar-based representations are more robust and transferable.

\noindent$\bullet$~\textbf{Backbone Transferability.}
To investigate the backbone transferability, we train LeaPRR using similarities from a source base backbone and evaluated it on similarities of another target one. As shown by the impressive improvement on the overall metric (rSum) in Figure~\ref{fig:model_transferability}, LeaPRR exhibits superior transferability across almost all base ITR architectures. It is attributed to the robust pillar-based representation which is consistent across various architectures.  

\begin{table}[t]
	\centering
	\setlength{\abovecaptionskip}{0cm}
	\caption{Dataset transferability results achieved by LeaPRR on top of DIME$^*$~\cite{qu2021dynamic}. The transferability results and the overall performance improvement on rSum compared with base backbones are marked in bold and {\color{mygreen}green}, respectively.}
	\vspace{1em}
	\label{tab:dataset_transferability}
	{      
		\resizebox{0.25\textwidth}{!}
		{
			\setlength\tabcolsep{2pt}
			\renewcommand\arraystretch{1.2}
            \begin{tabular}{l||c|c}
            \hline \thickhline 
            \diagbox{Source}{Target} &  Flickr30K &  MS-COCO  \\
            \hline \hline
            Base & 520.0 & 438.8  \\
            Flickr30K & \reshll{535.9}{15.9} & \reshl{446.8}{11.0}  \\
            MS-COCO & \reshl{529.1}{9.1} & \reshll{448.5}{12.7}  \\
            \hline
            \end{tabular}
		}
	}
    \vspace{-3ex}
\end{table}

\noindent$\bullet$~\textbf{Dataset Transferability.}
To further delve into the transferability across datasets, we also carried out corresponding experiments, the results of which are reported in Table~\ref{tab:dataset_transferability}. The comparison and performance improvement over the base backbone illustrates the outstanding dataset transferability of LeaPRR. It is worth noting that the improvement for Flickr30K $\rightarrow$ MS-COCO is highly comparable to that of learning on MS-COCO, specifically 11.0 vs. 12.7, although the scale of Flickr30K is only a quarter of that of MS-COCO. We ascribe the strong transferability to the extensive soundness of the pillar-based representation, \ie, representing an entity with its top-ranked neighbors, as discussed in Section~\ref{sec:pillar_rep}.

\section{Conclusion and Future Work}
In this paper, we proposed to model the higher-order neighbor relations for image-text retrieval in the re-ranking paradigm. To this end, we first delved into the limitations of prior re-ranking methods in the context of multi-modality and proposed four challenges for image-text retrieval re-ranking. To tackle the four issues, we then reformulated multi-modal re-ranking, constructed a new pillar space with top-ranked neighbors, and proposed a learning-based framework, which can fit in with one-tower and two-tower frameworks and flexibly explore complex neighbor relations among entities. Comprehensive experiments demonstrate the effectiveness, superiority, generalization, and transferability of the proposed method.

In the future, to further promote re-ranking performance, we plan to introduce abundant multimodal external knowledge for the modeling of the higher-order neighbor relations. Besides, inspired by the proposed re-ranking paradigm, we plan to explore the transferability between different downstream retrieval tasks from a broader perspective. 

\bibliographystyle{ACM-Reference-Format}
\balance
\bibliography{sample-base}


\end{document}